\documentclass[lettersize,journal]{IEEEtran}
\usepackage{amsmath,amsfonts}
\usepackage{algorithmic}
\usepackage{array}
\usepackage[caption=false,font=normalsize,labelfont=sf,textfont=sf]{subfig}
\usepackage{textcomp}
\usepackage{stfloats}
\usepackage{url}
\usepackage{verbatim}
\usepackage{graphicx}
\usepackage{cite}

\usepackage{multirow}
\usepackage{amssymb}
\usepackage{pifont}
\usepackage{graphicx}
\usepackage{mathtools,amsthm}

\usepackage{bm}

\begin{document}

\title{Generative AI for Synthetic Data Generation: Methods, Challenges and the Future}

\author{Xu Guo,~\IEEEmembership{Member,~IEEE}, and Yiqiang Chen,~\IEEEmembership{Senior Member,~IEEE}
\thanks{Xu Guo is with Nanyang Technological University (NTU), Singapore and Yiqiang Chen is with Institute of Computing
Technology, Chinese Academy of Sciences.}
\thanks{Emails: xu008@e.ntu.edu.sg}}



\maketitle

\begin{abstract}
The recent surge in research focused on generating synthetic data from large language models (LLMs), especially for scenarios with limited data availability, marks a notable shift in Generative Artificial Intelligence (AI). Their ability to perform comparably to real-world data positions this approach as a compelling solution to low-resource challenges. This paper delves into advanced technologies that leverage these gigantic LLMs for the generation of task-specific training data. We outline methodologies, evaluation techniques, and practical applications, discuss the current limitations, and suggest potential pathways for future research.
\end{abstract}


\section{Introduction}
\label{sect:intro}  
The introduction of Transformer \cite{vaswani2023attention} in 2017, followed by groundbreaking LLMs like OpenAI's GPT \cite{brown2020language} and Google's BERT \cite{devlin-etal-2019-bert}, marked the beginning of a new era in language understanding and generation. More recently, generative LLMs (e.g., GPT-3\cite{kojima2023large}, LlaMa\cite{touvron2023llama} and ChatGPT\cite{chatgpt}) have propelled this evolution to unprecedented heights, seamlessly converging with Generative AI and heralding a fresh era in the realm of synthetic data generation\cite{meng2023tuning,meng2022generating,zerogen,gao2023selfguided,ye-etal-2022-progen,regen,chen-etal-2023-mixture}. 

The origins of Generative AI can be traced back to pivotal models such as Generative Adversarial Networks\cite{goodfellow2014generative} (GANs) and Variational Autoencoders\cite{kingma2022autoencoding} (VAEs), which demonstrated the ability to generate realistic images and signals\cite{wu2020logan}. However, it wasn't until the advent of LLMs in recent years that Generative AI truly began to flourish. These LLMs, trained on vast datasets, showcased an unprecedented ability to produce coherent and contextually relevant text, pushing the boundaries of what AI could achieve in language-related tasks. The convergence of Generative AI and LLMs in the realm of synthetic data creation represents not merely a technological advancement, but a profound paradigm shift in our approach to data creation and the training of AI models.

\vspace{0.2cm}
\noindent\textbf{Why do we need synthetic data?} The necessity for synthetic data arises from the inherent limitations of general-purpose Large Language Models (LLMs) in specialized and private domains, despite their significant achievements across various benchmarks. For instance, ClinicalBERT\cite{huang2019clinicalbert}, adapted from BERT through pre-training on clinical texts, demonstrates superior performance in predicting hospital readmissions compared to the original BERT\cite{devlin2018bert}, which was trained on Wikipedia and BookCorpus\cite{zhu2015aligning} text data. This highlights a crucial challenge: specialized domains often rely on domain-specific data that is not readily available or open to the public, thereby underscoring the importance of synthetic data in bridging these gaps.

\vspace{0.2cm}
\noindent\textbf{Synergy between LLMs and synthetic data generation.}
Large Language Models (LLMs) for synthetic data generation marks a significant frontier in the field of AI. LLMs, such as ChatGPT, have revolutionized our approach to understanding and generating human-like text, providing a mechanism to create rich, contextually relevant synthetic data on an unprecedented scale. This synergy is pivotal in addressing data scarcity and privacy concerns, particularly in domains where real data is either limited or sensitive. By generating text that closely mirrors human language, LLMs facilitate the creation of robust, varied datasets necessary for training and refining AI models across various applications, from healthcare\cite{peng2023study}, eduction\cite{moore2023empowering} to business management\cite{rane2023role}. Moreover, this collaboration opens new avenues for ethical AI development, allowing researchers to bypass the biases and ethical dilemmas often inherent in real-world datasets. The integration of LLMs in synthetic data generation not only pushes the boundaries of what's achievable in AI but also ensures a more responsible and inclusive approach to AI development, aligning with evolving ethical standards and societal needs.

\vspace{0.2cm}
\noindent\textbf{Other related survey papers.} Comprehensive surveys for Generative AI and LLMs exist, each revisits related works from a different perspective: Generative AI surveys provide a holistic view of this area starting from Generative Adversarial Networks (GANs) to ChatGPT \cite{cao2023comprehensive} and models developed for synthetic data generation in the past decade \cite{bauer2024comprehensive}, with a special focus on text-to-image \cite{zhang2023text} or text-to-speech \cite{zhang2023survey} generation as well as practical applications in Education \cite{baidoo2023education} and Healthcare \cite{yu2023leveraging}; Surveys for LLMs provide systematic categorization \cite{qiu2020pre} for NLP tasks \cite{min2023recent} and methods to adapt these LLMs to specific domains \cite{10356_167965} through model optimization and personalization perspectives \cite{guo2022domain}. Surveys on LLMs for text generation \cite{li2022pretrained} focus on developing generative LLMs including model architecture choices and training techniques and do not contain gigantic LLMs released in the past two years. Unlike these survey papers, this paper mainly focuses on recent technologies that employ generative LLMs \textit{without training} them for synthetic training data generation and elicit their potential impact on practical adoption.

\vspace{0.2cm}
\noindent\textbf{Outline of this paper.} The following of this paper is organized as follows. Section \ref{sec:mthods} introduces recent methods for generating synthetic data from LLMs. Specifically, we summarize prompt engineering techniques that are particularly designed for probing LLMs to obtain desired data in sub-section \ref{sec:methods-prompt} while in sub-section \ref{sec:methods-adaptation}, we talk about how to employ parameter-efficient methods to adapt LLMs for generating task-related data; In sub-sections \ref{sec:methods-quality} and \ref{sec:methods-training} we introduce methods that can measure the quality of the synthetic dataset and how to effectively make use of the data for training. Section \ref{sec:application} details the application of synthetic data, focusing on its utilization in low-resource tasks in Sub-Section \ref{sec:application-low} and practical deployment scenarios in Sub-Section \ref{sec:application-fast}. Additionally, Sub-Section \ref{sec:application-medical} provides a specific case study on the use of synthetic data within medical domains. Finally, in Section \ref{sec:challenges}, we underscore some prominent challenges in synthetic data and discuss potential avenues for future research.
\begin{figure*}[t]
    \centering
    \includegraphics[width=0.8\textwidth]{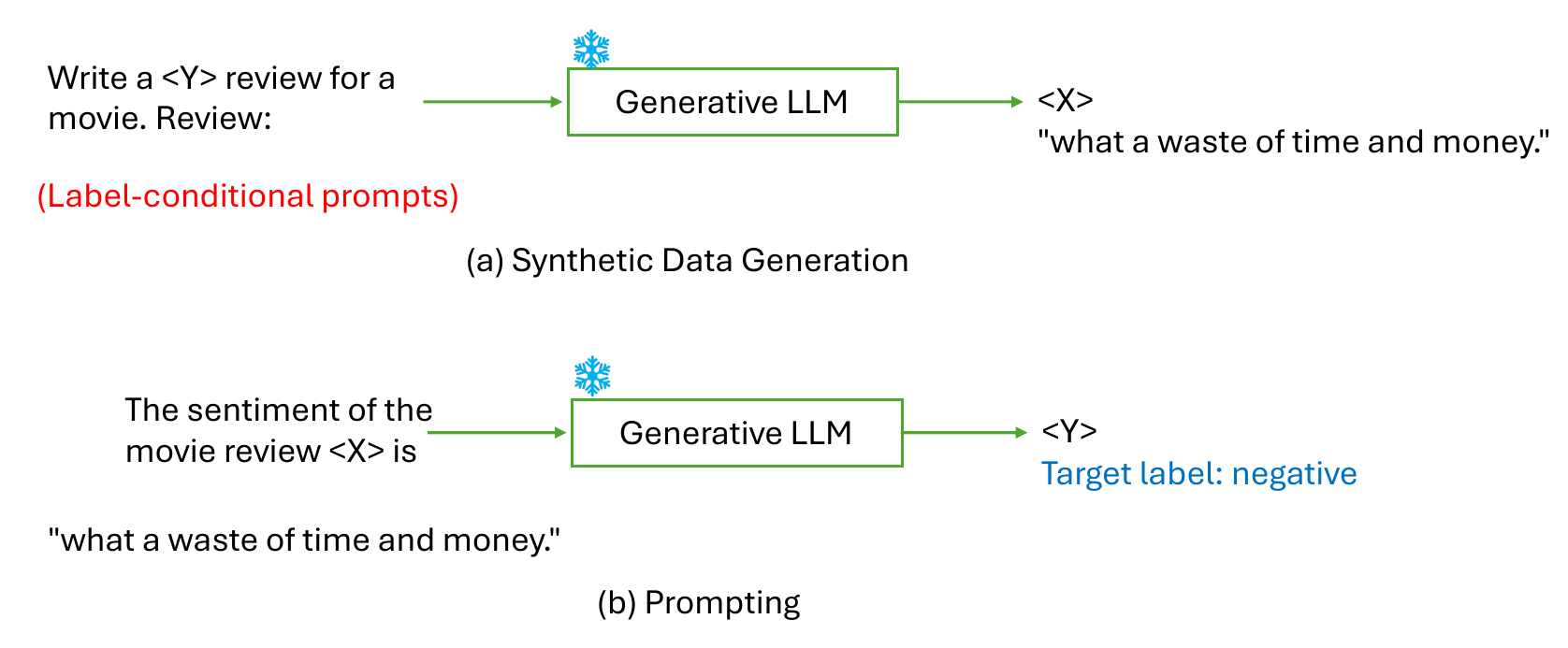}
    \caption{A general comparison between using LLMs for label-specific synthetic data generation (a) and label words prediction (b). In both cases, the LLMs are frozen and a task-related prompt is provided to condition the LLMs for task adaptation. $\langle X\rangle$ represents the text data and $\langle Y\rangle$ represents the label words. }
    \label{fig:workflow}
\end{figure*}

\section{Generating synthetic training data from LLMs}\label{sec:mthods}

Figure \ref{fig:workflow} shows the major difference between using generative LLMs for synthetic data generation and the predominant $\mathrm{Prompting}$ technique \cite{brown2020language,raffel2020exploring} that directly applies LLMs for label prediction. In short, $\mathrm{Prompting}$ requires deploying the LLM model in practice to predict the label words $\langle Y\rangle$ (e.g., negative) from the input text data $\langle X\rangle$ with additional constraints from the prompt, e.g., ``the sentiment of the movie review" indicates that the context is a movie review and the label shall describe its sentiment. On the contrary, synthetic data generation requires LLMs to generate text data $\langle X\rangle$ based on label-conditional prompts. It is the synthetic data distilled from LLMs rather than the LLMs themselves that will be applied in downstream applications, enabling more diverse and unlimited use cases based on synthetic data. Table \ref{tab:gen_methods} lists the newly emerging methods for generating task-specific training data from LLMs proposed in the past two years.

\begin{table*}[t]
    \centering
    \small
    \begin{tabular}{c|c|c|c}\hline
        Method & Generator & Classifier & Benchmark  \\\hline
        ZeroGen \cite{zerogen} & GPT2-XL \cite{radford2019language} & LSTM\cite{hochreiter1997long} & SST-2\cite{socher-etal-2013-recursive}, IMDb\cite{maas-etal-2011-learning}, QNLI\cite{squad}\\
        && DistilBERT \cite{sanh2020distilbert}& RTE\cite{dagan2005pascal}, SQuAD\cite{squad}\\
        &&& AdversarialQA\cite{bartolo-etal-2020-beat} \\\hline
        
        ZeroGen$^{+}$ \cite{gao2023selfguided} & GPT2-XL\cite{radford2019language} & LSTM\cite{hochreiter1997long} & IMDb\cite{maas-etal-2011-learning}, SST-2\cite{socher-etal-2013-recursive}, Amazon\cite{mcauley2013hidden} \\
        &&DistilBERT \cite{sanh2020distilbert}& Rotten Tomatoes\cite{pang-lee-2005-seeing}, Yelp\cite{zhang2015character} \\
        &&& Subj\cite{pang-lee-2004-sentimental}, AGNews\cite{zhang2015character}, DBpedia\cite{zhang2015character} \\\hline
        
        SuperGen \cite{meng2022generating} & CTRL\cite{keskar2019ctrl} & COCO-LM\cite{meng2021coco} & GLUE\cite{wang-etal-2018-glue} \\
        && RoBERTa\cite{liu2019roberta} & \\
        &&GPT-2\cite{radford2019language} &\\\hline
        
        FewGen \cite{meng2023tuning}& CTRL\cite{keskar2019ctrl} & RoBERTa\cite{liu2019roberta} & GLUE\cite{wang-etal-2018-glue} \\\hline
        
        ReGen \cite{regen} & Condenser\cite{condenser} & RoBERTa\cite{liu2019roberta} & AGNews\cite{zhang2015character},DBpedia\cite{zhang2015character}, MR\cite{pang-lee-2005-seeing} \\
        &&& NYT\cite{Meng_Shen_Zhang_Han_2019}, Yahoo\cite{zhang2015character}, Amazon\cite{mcauley2013hidden}\\
        &&& Yelp\cite{zhang2015character}, SST-2\cite{socher-etal-2013-recursive}, IMDb\cite{maas-etal-2011-learning} \\\hline
        
        ProGen \cite{ye-etal-2022-progen} & GPT2-XL\cite{radford2019language} & LSTM\cite{hochreiter1997long} & SST-2\cite{socher-etal-2013-recursive}, IMDb\cite{maas-etal-2011-learning}, Elec\cite{mcauley2013hidden}\\
        &&DistilBERT \cite{sanh2020distilbert}& Rotten Tomatoes\cite{pang-lee-2005-seeing}, Yelp\cite{zhang2015character}\\\hline
        
        AttrPrompt \cite{yu2023large} & ChatGPT\cite{chatgpt} & BERT\cite{devlin-etal-2019-bert} & NYT\cite{Meng_Shen_Zhang_Han_2019}, Amazon\cite{biographies} \\
         &  &  DistilBERT \cite{sanh2020distilbert} & Reddit\cite{geigle2021tweac}, StackExchange\cite{geigle2021tweac} \\\hline
        MixPrompt \cite{chen-etal-2023-mixture} & FLAN-T5 XXL \cite{chung2022scaling} & GODEL \cite{peng2022godel} & NLU++\cite{casanueva},TOPv2\cite{chen-etal-2020-low} \\
         &  &  &  CrossNER \cite{Liu2021} \\\hline
        
    \end{tabular}
    \vspace{0.2cm}
    \caption{Data generation methods. Generator refers to LLMs that are used for synthetic data generation. Classifier refers to small-scale models that are trained on the synthetic data. These methods are limited to NLP models and tasks.}
    \label{tab:gen_methods}
\end{table*}

\subsection{Prompt engineering}\label{sec:methods-prompt}
Designing an informative prompt is the key to effective data generation with LLMs. A simple and straightforward approach is to embed the label information in the prompt to refrain LLMs from generating label-agnostic data as described in Figure \ref{fig:workflow} (a). However, due to the limited number of words in labels and the limited task information in the prompt, the data generated by LLMs still can be unrelated to the task and lack diversity, limiting the size of the synthetic dataset that can be generated from the same LLM. As such, more advanced prompt engineering techniques are expected to circumvent the limitations of traditional ones.

\vspace{0.2cm}
\noindent\textbf{Attribute-controlled prompt.} A clear definition for a specific task can be obtained by specifying a set of attributes. Take News classification as an example, one piece of News article can differ from another by providing the details of $\mathrm{location}$, $\mathrm{topic}$, $\mathrm{text}$ $\mathrm{genre}$ and so on. Inspired by this, MSP \cite{chen-etal-2023-mixture} employs a mixture of attributes in the prompt template to obtain desired synthetic data. In AttrPrompt \cite{yu2023large}, authors show that such attribute-specific prompts can be directly extracted from ChatGPT and then applied to query ChatGPT for generating attribute-specific data. By expanding the simple class-conditional prompt with more attribute constraints, we can gather more diverse synthetic data from LLMs while ensuring relevance to the given task.

\vspace{0.2cm}
\noindent\textbf{Verbalizer.} The verbalizer technique was originally proposed to enhance $\mathrm{Prompting}$ performance, where the target label words are expanded with their neighbouring words that hold the same semantic meanings \cite{cui-etal-2022-prototypical,hu2022knowledgeable}. This strategy can be directly utilized to promote diverse data generation by expanding the class-conditional prompt into a set of semantically similar prompts. Besides, the verbalizer values can be extracted from LLMs themselves. For example, MetaPrompt \cite{reynolds2021prompt} first obtains an expanded prompt from ChatGPT and further applies the enriched prompt to prompt LLMs for data generation.

\subsection{Parameter-efficient task adaptation}\label{sec:methods-adaptation}
Parameter-efficient approaches in the era of LLMs generally refer to the tuning methods that only tune a small set of an LLM's parameters (e.g., bias terms \cite{ben-zaken-etal-2022-bitfit}, embeddings or last layer) or an extra set of parameters that are inserted to LLMs (e.g., Adapters \cite{houlsby19a,liu-etal-2023-intemats}, Prompt Tuning \cite{lester-etal-2021-power,guo-etal-2022-improving}, Prefix Tuning \cite{li-liang-2021-prefix} and LoRA \cite{hu2022lora}). In the tuning process, the parameters of the LLM backbone are not updated and only the small set of trainable parameters are learned on task-specific datasets to achieve domain adaptation. More parameter-efficient methods can be found in the survey \cite{ding2022delta}. The advantage of parameter-efficient methods is that they grasp new task information while retaining powerful pre-trained knowledge.

To enable a general LLM to generate data for a specific task style, one promising approach is to aggregate a few-shot dataset (e.g., eight instances per class) and perform parameter-efficient adaptation for the LLM \cite{guo-etal-2022-improving}. The method, FewGen \cite{meng2023tuning}, demonstrates that by tuning a few set of prefix vectors prepended to the CTRL model (1.6 Billion parameters) on few-shot datasets, the PrefixCTRL can generate more task-related training data. Similarly, MSP \cite{chen-etal-2023-mixture} trains a set of soft prompt embeddings on few-shot task-specific training data and then applies the trained soft prompts to condition the FLAN-T5 \cite{chung2022scaling} (T5\cite{raffel2020exploring} further trained on instruction tuning datasets) for text generation. Compared with zero-shot generation, a small budget for few-shot task data can allow the general-purpose LLMs to quickly adapt to the target task under the parameter-efficient learning paradigm.


\subsection{Measuring data quality}\label{sec:methods-quality}
The quality of synthetic data is often measured by quantitative metrics. In ZeroGen \cite{zerogen}, authors measured the quality of the generated data from three perspectives: $\mathrm{diversity}$, $\mathrm{correctnes}$, and $\mathrm{naturalness}$.  $\mathrm{Diversity}$ measures the difference between a chunk of text and another in the generated instances. For example, 4-$\mathrm{gram}$ Self-BLEU computes BLEU scores on every four consecutive tokens in the generated texts. $\mathrm{Correctnes}$ measures whether the data instance is related to the given label. Existing approaches for measuring $\mathrm{correctnes}$ can be divided into two categories: automatic evaluation and human evaluation. Automatic evaluation methods train a model (e.g., RoBERTa-large) on the oracle training dataset in a fully-supervised full-model fine-tuning manner, and then apply the model to calculate the percentage of correctly predicted samples on the synthetic dataset. Human evaluation requires the availability of human annotators who will be assigned a random subset of the synthetic dataset and asked to judge whether the content is related to the label. $\mathrm{Naturalness}$ measurement requires human evaluators who can assess whether the generated text is fluent and similar to human-written texts by selecting a score from a given range.

To obtain high-quality synthetic data, ProGen \cite{ye-etal-2022-progen} proposes to incorporate a quality estimation module in the data generation pipeline, where the firstly generated synthetic data are evaluated by a task-specific model that was trained on oracle data in advance. Then, the most influential synthetic samples are selected as in-context examples to prompt GPT2-XL \cite{radford2019language} to generate a new set of synthetic data.

\subsection{Training with synthetic data} \label{sec:methods-training}
In the process of training with synthetic data generated from LLMs, challenges such as inherent biases and hallucinations in the LLMs can introduce noise into the dataset, despite meticulous prompt design and supervised training. To mitigate these issues, the implementation of regularization techniques is crucial for stabilizing training with noisy datasets. Innovations like ZeroGen$^{+}$ \cite{gao2023selfguided} suggest the use of a small weight network trained through bilevel optimization to autonomously determine sample weights. Additionally, FewGen \cite{meng2023tuning} incorporates a self-supervised training approach using temporal ensembling \cite{laine2016temporal}. This method has been shown to offer superior performance enhancements compared to label smoothing \cite{müller2020does} when training downstream classifiers on synthetic data, highlighting its effectiveness in dealing with the unique challenges posed by synthetic datasets. Other techniques such as gradual annealing \cite{du2023training} also demonstrates to be effective in enhancing the learning performance on synthetic data.

\section{Applications}\label{sec:application}
Synthetic data generated from LLMs can be used in a wide range of applications. In this section, we first introduce how to solve the long-standing low-resource and long-tail problems with synthetic data and its use cases for fast inference and deployment. Then, we present two practical examples of applying synthetic data in medical and education scenarios. 

\subsection{Low-resource and long-tail problems}\label{sec:application-low}
Low-resource problems are generally trapped by the lack of sufficient data and in some cases particularly impacted by long-tail classes in practice \cite{tiong2023improving}. Traditional research has predominantly leveraged transfer learning techniques \cite{guo-etal-2021-latent, guo-etal-2022-improving} to enhance performance in low-resource settings. Yet, these methods hinge on the availability of relevant source-domain datasets, which may not always be accessible. The impressive generative capabilities of LLMs and the production of highly realistic synthetic data signal a significant potential to reshape the traditional landscape of low-resource and long-tail problems. 

A primary challenge in merging the research directions of synthetic data generation and low-resource learning tasks is navigating the distribution disparity between real and synthetic data, as well as optimizing the use of synthetic data in training scenarios. Noteworthy approaches to address these issues include the application of regularization techniques. For instance, FewGen employs temporal ensembling \cite{meng2023tuning}, and CAMEL utilizes gradual learning \cite{du2023training}. Additionally, innovative data selection techniques, as explored in Du et al. (2023) \cite{du2023training2}, offer valuable insights. These methods are instrumental in harnessing the full potential of synthetic data to enhance learning performance, particularly in environments where real data is limited or imbalanced.

\subsection{Fast inference and lightweight deployment}\label{sec:application-fast}
Finetuning pre-trained language models on downstream tasks has been the predominant approach starting from the release of BERT \cite{devlin2018bert}. However, the growing size of these language models, while enhancing performance, imposes practical burdens on organizations requiring swift inference and prompt responses. The shift towards synthetic data generation opens up a realm of possibilities for downstream applications. By generating a curated synthetic dataset, it becomes feasible to train smaller, less complex models, as demonstrated in \cite{zerogen,gao2023selfguided,ye-etal-2022-progen}. This approach not only facilitates easier deployment but also ensures faster inference, addressing the critical need for efficiency in real-world applications.

\subsection{Medical Scenarios}\label{sec:application-medical}

The medical domain presents unique challenges due to the confidential nature of patient data and the relative scarcity of medical data compared to the abundance of information available online. The use of LLMs and multi-modal LLMs has shown promising potential in medical domains such as dental diagnosis \cite{huang2023chatgpt}, radiograph analysis \cite{packhauser2023generation}, and so on \cite{singhal2023large,thirunavukarasu2023large}. The exceptional data comprehension and generation capabilities of LLMs position synthetic data generation as an especially promising research avenue in the medical domain.

\noindent\textbf{Data augmentation.} Synthetic data generation can help some medical tasks that lack sufficient data to train a strong predictive model. For instance, studies in \cite{packhauser2023generation} demonstrated that augmenting real datasets with synthetic chest radiograph images generated by latent diffusion models\cite{rombach2022highresolution} can enhance classification performance. In medical language processing, Tang et al. (2023) \cite{tang2023does} demonstrated that tailored prompts provided to ChatGPT can yield task-specific synthetic data, significantly boosting the performance in tasks like biological named entity recognition and relation extraction.  Additionally, GatorTronGPT, as explored in Peng et al. (2023) \cite{peng2023study}, which involved training GPT-3 from scratch on a dataset amalgamating 277-billion words from English and clinical texts, exhibited remarkable proficiency in generating synthetic clinical text. This data surpassed real data in performance across various biomedical tasks, including relation extraction and question answering, showcasing the potential of synthetic data in transforming medical AI applications.

\noindent\textbf{Missing value imputation.} Medical data can be sparse in that patients may take different or do not take some examinations, leading to imbalanced attributes. Missing value imputation (MVI) methods are helpful in enhancing the density of medical attribute values \cite{liu2023handling}. Traditional MVI approaches typically involve random sampling from specified value ranges, as noted in Luo et al. (2022) \cite{luo2022missing}, essentially serving as a form of random data augmentation for certain attributes. With the advent of multi-modal LLMs, Ozbey et al. (2023) \cite{ozbey2023unsupervised} demonstrate that in cross-modality translation tasks, missing images under specific attributes can be effectively imputed using synthetic images generated from diffusion models. Such synthetic data, compared to traditional random imputation methods, offer more diverse information, thereby helping to mitigate the issue of overfitting in attributes with limited data.


\section{Challenges with Synthetic Data and Future Directions}\label{sec:challenges}
Many domains suffer from a lack of quality data, especially when it comes to rare events or minority classes. LLMs can augment existing datasets, creating balanced and comprehensive data sets that improve the training and performance of machine learning models. In this section, we highlight some challenges in the creation and use of synthetic data and discuss promising research directions. 

\subsection{Overcoming Data Limitations}
Synthetic data generated from LLMs inherently faces several data limitations that must be acknowledged and addressed. 

\noindent\textbf{Correctness and Diversity.} In Section \ref{sec:mthods}, we summarized existing approaches for monitoring the data quality and promoting data diversity in generation. They demonstrated effectiveness but do not entirely solved the problem. The challenge of ensuring the quality and accuracy of the generated data still remains profound. As an inherent nature, LLMs may inadvertently propagate inaccuracies or biases present in their pre-training data \cite{liang2021towards,kotek2023gender}, leading to outputs that may not always align with factual or unbiased information. Additionally, the intra-class and inter-class data diversity and domain representativeness are a concern, especially in specialized or niche domains.

\noindent\textbf{Hallucination.} Synthetic data generated by Large Language Models (LLMs) can sometimes be not only inaccurate but completely fictitious or disconnected from reality, a phenomenon often referred to as "hallucination" \cite{ji2023survey,zhang2023siren}. For instance, image generation based on specific captions can result in outputs with unrealistic features, such as a soldier depicted with three hands, as noted in the studies \cite{du2023training} for cross-modality generation. This hallucination issue is frequently linked to the quality of the training data, particularly if it contains inaccuracies that the LLM then overfits during the pre-training phase. The challenge is compounded due to the difficulty of either fine-tuning LLMs or modifying their pre-training data. Therefore, there's a pressing need to develop new, more effective strategies to detect and address hallucination \cite{xu2023understanding} in the context of synthetic data generation, ensuring the reliability and authenticity of the output.

\subsection{Data privacy and ethical concerns}
While synthetic data offers a way to leverage the power of AI without compromising individual privacy\cite{fang2023synthaspoof}, the ethical implications of using synthetic data, particularly in sensitive domains, raise questions about privacy and consent, as the boundaries between real and synthetic data blur. Research in \cite{carlini2021extracting} demonstrates that it is possible to extract specific information from the datasets used in training LLMs. Consequently, there exists a risk that synthetic data generation might inadvertently reveal elements of the underlying training data \cite{mccoy2023much}, some of which might be subject to licensing agreements. This scenario poses not only privacy issues but also potential financial implications for users, highlighting the need for careful management and consideration in the use and dissemination of synthetic data generated by LLMs.

Moreover, uploading data to LLM APIs also remains a data privacy concern. For instance, employing LLMs in clinical text mining poses significant privacy risks related to uploading patient information to LLM APIs \cite{tang2023does}. This challenge necessitates a careful balance between leveraging the benefits of AI and respecting the confidentiality and privacy of individuals, particularly in healthcare and other sensitive areas. Addressing these concerns requires not just technological solutions, but also robust policy frameworks and ethical guidelines to ensure responsible use of synthetic data and AI technologies. 

\section{Conclusion}
This paper reviews recent research on utilizing generative LLMs for synthetic data generation. With a focus on gigantic LLMs which are fixed for inference, we elicit the complexities of producing high-quality and diverse synthetic data and present some recent effective strategies to navigate these challenges, including attribute-controlled prompt engineering and verbalization strategies. Additionally, we also introduce some practical training techniques for training downstream models on the synthetic data presuming the data quality is inadequate. Then, we introduce some application scenarios for the use of synthetic data generation, extending from general low-resource issues to more specialized medical contexts. Finally, we conclude by spotlighting the significant ongoing challenges in the realm of synthetic data and proposing potential avenues for future research in this evolving field.

\section*{Acknowledgments}
Xu Guo thanks the Wallenberg-NTU Presidential Postdoctoral Fellowship.

\bibliography{custom}
\bibliographystyle{IEEEtran}

\vspace{-3em}












\newpage

 




\vfill

\end{document}